\documentclass{article}

\usepackage{microtype}
\usepackage{graphicx}
\usepackage{subfigure}
\usepackage{booktabs} 

\usepackage{hyperref}

\usepackage[accepted]{icml2024}

\usepackage{amsmath}
\usepackage{amssymb}
\usepackage{mathtools}
\usepackage{amsthm}

\usepackage[capitalize,noabbrev]{cleveref}

\theoremstyle{plain}

\theoremstyle{definition}

\theoremstyle{remark}

\usepackage{color}
\usepackage{subcaption}
\usepackage{multirow}

\def\I{{\mathbf{I}}}
\def\rvy{{\mathbf{u}}}
\def\rvs{{\mathbf{y}}}
\def\rvx{{\mathbf{x}}}
\def\rvz{{\mathbf{z}}}
\def\rvzl{{\mathbf{\tilde{z}}}}
\def\rvyl{{\mathbf{\tilde{u}}}}

\def\N{{\mathcal{N}}}
\def\E{{\mathbb{E}}}
\def\KL{{\mathrm{KL}}}
\def\p{p_{\rm data}}
\graphicspath{{compressed_images/} }
\usepackage{wrapfig}
\newcommand\blfootnote[1]{%
  \begingroup
  \renewcommand\thefootnote{}\footnote{#1}%
  \addtocounter{footnote}{-1}%
  \endgroup
}
\usepackage[textsize=tiny]{todonotes}

\begin{document}

\twocolumn[
\icmltitle{Learning Latent Space Hierarchical EBM Diffusion Models}




\begin{icmlauthorlist}
\icmlauthor{Jiali Cui}{stevens}
\icmlauthor{Tian Han}{stevens}
\end{icmlauthorlist}

\icmlaffiliation{stevens}{Department of Computer Science, Stevens Institute of Technology}

\icmlcorrespondingauthor{Tian Han}{than6@stevens.edu}

\icmlkeywords{Machine Learning, ICML}

\vskip 0.3in
]



\printAffiliationsAndNotice{}  

\begin{abstract}
This work studies the learning problem of the energy-based prior model and the multi-layer generator model. The multi-layer generator model, which contains multiple layers of latent variables organized in a top-down hierarchical structure, typically assumes the Gaussian prior model. Such a prior model can be limited in modelling expressivity, which results in a gap between the generator posterior and the prior model, known as the prior hole problem. Recent works have explored learning the energy-based (EBM) prior model as a second-stage, complementary model to bridge the gap. However, the EBM defined on a multi-layer latent space can be highly multi-modal, which makes sampling from such marginal EBM prior challenging in practice, resulting in ineffectively learned EBM. To tackle the challenge, we propose to leverage the diffusion probabilistic scheme to mitigate the burden of EBM sampling and thus facilitate EBM learning. Our extensive experiments demonstrate a superior performance of our diffusion-learned EBM prior on various challenging tasks. 
\end{abstract}

\section{Introduction}
The hierarchical generative model with multiple layers of latent variables (a.k.a., \textit{multi-layer generator model}) has made significant progress in learning complex data distribution \cite{NIPS2016_6ae07dcb,vahdat2020nvae} and has garnered particular interest for its top-down hierarchical structure, where multiple layers of latent variables that are organized from the top to the bottom layers tend to capture levels of (hierarchical) data representations, with high-level semantic representations captured by the latent variables at the top layers and low-level detail representations by those at the bottom layers \cite{maaloe2019biva,child2020very}. Learning such hierarchical representation can be essential and crucial to various downstream applications \cite{havtorn2021hierarchical,nijkamp2020learning}. However, such multi-layer generator models typically assume the Gaussian prior model, which can be limited in statistical expressivity by primarily focusing on the inter-layer relation between layers of latent variables while largely ignoring the intra-layer relation between latent units within each layer \cite{Cui_2023_CVPR,Cui_2023_ICCV}. This may result in the \textit{prior hole problem} \cite{rosca2018distribution,hoffman2016elbo,takahashi2019variational,bauer2019resampled} where the non-expressive Gaussian prior fails to match the aggregated generator posterior. 

Recent studies \cite{aneja2021contrastive,Cui_2023_CVPR} have investigated the utilization of the energy-based (EBM) prior model as a complementary model to address this limitation. The EBM prior is typically trained with a fixed generator model (referred to as the \textit{Two-Stage} learning scheme) to tilt the non-expressive Gaussian prior to match the generator posterior. However, learning a \textit{single} (marginal) EBM is challenging because the generator posterior is often multi-modal, and more importantly, the Markov Chain Monte Carlo (MCMC) sampling required to maximize the marginal EBM likelihood can be difficult, as multiple layers of latent variables are interwoven and require exploration at different latent scales. In addition, MCMC sampling, such as Langevin dynamics, usually starts from a noise-initialized point, which is hard to explore the energy landscape and mix between different local modes. Therefore, for multi-layer latent variables, EBM prior sampling may serve as the bottleneck for effective EBM learning, which still poses a challenge.

Inspired by recent diffusion probabilistic frameworks \cite{ho2020denoising,gao2020learning,zhu2023cdrl,du2023reduce}, we propose learning the EBM prior of multi-layer latent variables in a diffusion learning scheme. We construct a series of conditional EBMs prior to gradually matching the highly multi-modal generator posterior, with each conditional EBM prior only matching perturbed samples at each step. Compared to marginal EBM prior, such a conditional EBM prior can be less multi-modal, leading to more tractable conditional likelihood learning. For EBM sampling, the proposed conditional EBM prior can render a smoother energy landscape, which mitigates the burden of MCMC sampling and thus further facilitates effective EBM learning. However, for multi-layer latent variables, MCMC sampling needs to account for their different latent scales at different layers (i.e., the scales of latent variables at the top and bottom layers can be very different); moreover, directly perturbating the latent samples may destroy their inter-layer relation (i.e., conditional dependency formulated in the hierarchical structure). Therefore, we further employ a uni-scale $\rvyl$-space (see definition in Sec. \ref{sec-method-diffusion}) converted from the multi-scale latent space, which allows us to preserve the hierarchical dependency along the forward process while at the same time, further reducing the burden of MCMC sampling by traversing a uni-scale latent space. Our experiments demonstrate the effectiveness of the proposed method in various challenging tasks and show that our model is capable of generating high-quality samples and capturing hierarchical representations at different layers.

\textbf{Contribution:} 1) We develop a learning framework that incorporates the diffusion probabilistic scheme for learning the joint EBM prior for the multi-layer generator model; 2) To preserve hierarchical structures and enable more effective EBM sampling, we adopt a uni-scale space to further mitigate the burden of MCMC sampling; 3) We conduct various experiments to examine our model in generating high-quality samples and learning effective hierarchical representations.

\section{Preliminary}
\subsection{Multi-layer Latent Variable Model}\label{sec-bg-multi} 
Let $\rvx \in R^D$ be the high-dimensional observed example and $\rvz \in R^d$ be the low-dimensional latent variable. The multi-layer generator model contains multiple latent variables (i.e., $\rvz_1, \rvz_2, \dots, \rvz_L$) organized in a top-down hierarchical structure and can be specified as a joint distribution. We denote $\rvzl = (\rvz_1, \rvz_2, \dots, \rvz_L)$, then
\begin{equation}\label{gaussian-multi-joint}
\begin{aligned} 
     p_{\beta}(\rvx, \rvzl) =& p_{\beta_0}(\rvx|\rvzl)p_{\beta_{>0}}p(\rvzl)\;\;\;\;\text{where}\\
     p_{\beta_{>0}}(\rvzl) =& \prod_{i=1}^{L-1}p_{\beta_i}(\rvz_i|\rvz_{i+1})p(\rvz_{L})
\end{aligned}
\end{equation}
in which $p_{\beta_0}(\rvx|\rvzl)$ is the generation model that maps from the latent space to the data space, and $p_{\beta_{>0}}(\rvzl)$ is the prior model that factories consecutive layers of latent variables with conditional Gaussian distribution (i.e., $p_{\beta_i}(\rvz_i|\rvz_{i+1}) \sim \N (\mu_{\beta_i}(\rvz_{i+1}), \sigma^2_{\beta_i}(\rvz_{i+1}))$) parameterized by learnable parameter $\beta_i$. The $p(\rvz_{L}) \sim \N(0, \I_d)$ is assumed to be unit Gaussian at the top layer.

Learning such a hierarchical generative model can be achieved using maximum likelihood estimation (MLE) with the gradient estimated as $\nabla_\beta \E_{p_{\beta}(\rvzl|\rvx)}[\log p_\beta(\rvx, \rvzl)]$. For the generator posterior $p_{\beta}(\rvzl|\rvx)$, prior works utilize MCMC sampling to obtain approximated posterior samples \cite{nijkamp2020learning}, and \cite{NIPS2016_6ae07dcb,child2020very,maaloe2019biva} propose the variational learning that introduces a parameterized inference network (e.g., $q_\phi(\rvzl|\rvx)$) learned to approximate the generator posterior distribution.

However, such multi-layer generator models often fall short in generating high-quality image synthesis, as the Gaussian prior typically only focuses on the \textit{inter-layer} relation modelling while largely ignoring the \textit{intra-layer} relation modelling \cite{Cui_2023_CVPR}, resulting in the \textit{prior hole problem} with mismatch regions between the prior and aggregate posterior distribution \cite{dai2019diagnosing,ghosh2019variational}.

\subsection{Energy-based Prior Model.}\label{sec-bg-jebm} 
Another generative model, the energy-based model (EBM), is shown to be expressive in capturing the intra-layer relation and representing data uncertainty. In general, on data space $\rvx$, the EBM can be defined as
\begin{equation}\label{ebm}
    p_{\omega}(\rvx)=\frac{1}{\mathrm{Z}_{\omega}}\exp{[f_{\omega}(\rvx)]}
\end{equation}
where $\mathrm{Z}_{\omega}$ is the normalizing constant or partition function, $f_{\omega}(\rvx)$ is the energy function parameterized with $\omega$. 

Learning the EBM via MLE estimates the gradient as $\E_{\p(\rvx)}[\nabla_{\omega} f_{\omega}(\rvx) - \E_{p_{\omega}(\rvx)}[f_{\omega}(\rvx)]]$. For EBM samples from $p_{\omega}(\rvx)$, \cite{du2019implicit,du2020improved,nijkamp2019learning} adopt MCMC sampling such as Langevin dynamics (LD). In particular, it is applied as 
\begin{eqnarray}
\label{eq:ebm_langevin}
\rvx_{\tau+1} = \rvx_\tau + s \nabla_{\rvx_\tau}\log p(\rvx_\tau) + \sqrt{2s}U_\tau
\end{eqnarray}
where $\tau$ indexes the time step, $s$ is the step size and $U_\tau \sim \N(0, \I_D)$, and $\rvx_{\tau=0}$ is usually initialized from the Gaussian noise. However, in practice, it may take a long time to explore the energy landscape and mix between different local modes. To mitigate the burden of EBM sampling, recent advances have explored EBMs on low-dimensional latent space $p_{\omega}(\rvz)$ \cite{pang2020learning, xiao2022adaptive,yu2022latent}, but a single-layer $p_{\omega}(\rvz)$ prior model can still be limited in modelling compacity of the whole model.

\textbf{Two-stage complementary EBM prior.} Learning the EBM prior for multi-layer of latent variables $\rvzl = (\rvz_1, \rvz_2, \dots, \rvz_L)$ can be more expressive than single-layer latent variables, but jointly learning both the multi-layer generator model and EBM prior can be extremely inefficient, especially with a deep hierarchical structure involved \cite{vahdat2020nvae,child2020very}. This motivates a \textit{Two-Stage} learning scheme \cite{xiao2020vaebm,aneja2021contrastive,Cui_2023_CVPR} that learns the Gaussian prior generator model at the \textit{first stage} (see Sec. \ref{sec-bg-multi}) and then learns the EBM, as a complementary model at the \textit{second stage} with the fixed generator backbone. In our work, we adopt such a learning scheme for its efficiency. 

With multi-layer of latent variables, the NCP-VAE \cite{aneja2021contrastive} factories a conditional EBM prior 
\begin{equation}\label{cond-ebm-ncp}
    p_{\omega_i, \beta_i}(\rvz_i|\rvz_{i+1})=\frac{1}{\mathrm{Z}_{\omega_i, \beta_i}(\rvz_{i+1})}\exp{[f_{\omega_i}(\rvz_i)]}p_{\beta_i}(\rvz_i|\rvz_{i+1})\nonumber
\end{equation}
which aims to tilt the Gaussian prior toward the generator posterior distribution. The noise-contrastive estimation (NCE) is used for learning, which treats EBM as a \textit{classifier} and thus does not need MCMC approximation. However, the NCE scheme can render \textit{suboptimal learning} with a large gap between the two distributions \cite{xiao2022adaptive}, which exists between the generator posterior and the Gaussian prior \cite{arjovsky2017wasserstein,dai2019diagnosing}.

The recent work \cite{Cui_2023_CVPR} considers learning EBM prior via MLE by jointly modelling all layers of latent variables
\begin{equation}
\label{joint-ebm}
     p_{\omega, \beta_{>0}}(\rvzl) =\frac{1}{\mathrm{Z}_{\omega, \beta_{>0}}}\exp\left[F_{\omega}(\rvzl)\right]p_{\beta_{>0}}(\rvzl)
\end{equation}
where the energy function $F_{\omega}(\rvzl)=\sum_{i=1}^L f_{\omega_i}(\rvz_i)$. However, MCMC sampling for Eqn. \ref{joint-ebm} can be practically challenging as layers of $\rvz_i$ can have different scales (i.e., $\rvz_L \sim \N(0, \I_d)$ and $\rvz_1 \sim p_{\beta_{>0}}(\rvz_1) = \int p_{\beta_{>0}}(\rvz_1, \dots, \rvz_L) d\rvz_2, \dots, d\rvz_L$), which requires special designs for MCMC sampling to account for such variation.

Learning both the multi-layer EBM prior can be viewed to minimize the Kullback-Leibler (KL) divergence between the generator posterior distribution and the EBM prior, i.e., $\KL(p_{\theta}(\rvzl|\rvx) || p_{\omega, \beta_{>0}}(\rvzl))$, which is difficult due to the highly multi-modal generator posterior and the multi-scale latent space, resulting in ineffective MCMC sampling for EBM learning.

\section{Methodology}
Inspired by recent diffusion probabilistic methods that focus on learning a sequence of parameterized models to gradually match target data distribution, we study a probabilistic framework that can leverage such diffusion scheme with a sequence of \textit{conditional EBMs prior} for the \textit{multi-layer generator models}.

\subsection{Diffusion with Multi-layer Latent Variables}\label{sec-method-diffusion} 
\textbf{Attempt on $\rvzl$-space.} The diffusion probabilistic scheme assumes a sequence of perturbed samples $\rvz_{0:T}=(\rvz_0, \rvz_1, \dots, \rvz_T)$ for each diffusion step $t=0, 1, \dots, T$. In particular, the noisy sample $\rvzl_t$ is generated by a pre-defined Gaussian perturbation kernel as
\begin{equation}
\label{diff-forward-kernel-z}
    q(\rvzl_{t+1}|\rvzl_{t}) \sim \N(\alpha_{t+1}\rvzl_{t}, \sigma^2_{t+1}\mathbf{I}_{|d|})
\end{equation}
where $\alpha_{t+1}$ is typically set to be $\sqrt{1-\sigma^2_{t+1}}$ to ensure a spherical interpolation between samples and noise. 

However, directly employing Eqn. \ref{diff-forward-kernel-z} does not suit for multi-layer latent variables $\rvzl$, since it does not take into account the hierarchical structure between layers of latent variables. Their inter-layer relation is consequently \textit{destroyed} during the progress, i.e., each $\rvz_i$ becomes independently distributed as standard Gaussian noise at the final diffusion step. Our goal is to reach the Gaussian prior model $p_{\beta_{>0}}(\rvzl)$ (Eqn. \ref{gaussian-multi-joint}) at the final step, such that the reverse process can start from the Gaussian prior model to approximate the generator posterior distribution gradually.

\textbf{Toward $\rvyl$-space.} Instead of latent space, we formulate our diffusion model on \textit{$\rvyl$-space}. In particular, for multi-layer generator models, the Gaussian prior model $p_{\beta_{>0}}(\rvzl)$ is factorized to be the multiplication of consecutive layers of conditional Gaussian distribution $p_{\beta_i}(\rvz_i|\rvz_{i+1}) \sim \N (\mu_{\beta_i}(\rvz_{i+1}), \sigma^2_{\beta_i}(\rvz_{i+1}))$, which features the re-parametrization sampling, i.e., $\rvz_i = \mu_{\beta_i}(\rvz_{i+1}) + \sigma_{\beta_i}(\rvz_{i+1})\cdot \rvy_{i+1}$ where $\rvy_{i+1} \sim \N(0, \I_d)$. We denote such an invertible deterministic transformation function to be $T_{\beta_{>0}}$, i.e., $\rvzl = T_{\beta_{>0}}(\rvyl)$ and $\rvyl = T_{\beta_{>0}}^{-1}(\rvzl)$ (see App. \ref{appendix-derivation}), which allows us to adapt our diffusion model on $\rvyl$-space. The $\rvyl$-space enables the hierarchical structure to be maintained during the forward progress. When $q(\rvyl_T)\sim\N(0, \I_{|d|})$ becomes standard Gaussian noise after the forward diffusion process, the corresponding latent variables reach the desired Gaussian prior model at the final step.

\textbf{Forward on $\rvyl$-space.} The perturbation kernel is defined as
\begin{equation}
\label{diff-forward-kernel-y}
    q(\rvyl_{t+1}|\rvyl_{t}) \sim \N(\alpha_{t+1}\rvyl_{t}, \sigma^2_{t+1}\mathbf{I}_d)
\end{equation}
With a designed diffusion $\sigma$-schedule (e.g., signal-to-noise ratio, SNR), $\rvyl_{T}$ at final step $t=T$ becomes Gaussian noise and $q(\rvyl_{T}) = \N(0,\I_{|d|})$ is then the stationary distribution.

For the case of $t=0$, we obtain $\rvyl_0$ via transformation function $\rvyl_0 = T_{\beta_{>0}}^{-1}(\rvzl_0)$ with $\rvzl_0$ sampled from the generator posterior. For variational-based generator models (see Sec. \ref{sec-bg-multi}), $\rvzl_0$ can be inferred from the inference model, while we can also perform MCMC posterior sampling for $\rvzl_0$. For deep hierarchical structures, the inference model is preferred for efficiency. The forward trajectory then becomes

\begin{equation}
\label{diff-forward-traj-y}
    q_{\beta}(\rvyl_{0:T}|\rvx) = \prod_{t=0}^{T-1} q(\rvyl_{t+1}|\rvyl_{t})q_{\beta}(\rvyl_0|\rvx)
\end{equation}
where $q_{\beta}(\rvyl_0|\rvx)$ is the unknown target distribution. The goal is now to reverse from $\rvyl_{T}$ to $\rvyl_{0}$, which, in view of distributions, reverses from the Gaussian prior model $p_{\beta_{>0}}(\rvzl)$ to the generator posterior $p_{\beta}(\rvzl|\rvx)$. This strategy circumvents the problem of destroying hierarchical patterns during the forward process and thus satisfies our goal to bridge the gap between Gaussian prior model and the generator posterior. 

\subsection{Reverse with Multi-layer Latent Variables}
\textbf{$\rvyl$-space for marginal EBM prior.} The uni-scale $\rvyl$-space is adopted in works \cite{xiao2020vaebm,Cui_2023_CVPR} for EBM sampling. Specifically, sampling from the marginal EBM (Eqn. \ref{joint-ebm}) is equivalent to sampling from
\begin{equation}\label{marginal-y-ebm}
\begin{aligned}
 p_{\omega, \beta_{>0}}(\rvyl) = \frac{1}{\mathrm{Z}_{\omega, \beta_{>0}}}\exp \left[F_\omega(T_{\beta_{>0}}(\rvyl))\right]p_0(\rvyl)
\end{aligned}
\end{equation}
where $p_0(\rvyl)$ is standard Gaussian. The uni-scale $\rvyl$-space can make EBM sampling easier than the multi-scale latent space $\rvzl$. However, this is still challenging as it aims to match the highly multi-modal generator posterior and Gaussian prior with a single (marginal) EBM. 

To tackle this challenge, prior works \cite{gao2020learning,yu2022latent} leverage diffusion scheme and learn sequential conditional EBMs, which has seen some success in modelling the high-dimensional $\rvx$-space and single-layer $\rvz$-space. Inspired by their work, we propose to learn a sequence of conditional EBMs prior but focus on hierarchical generative models with $\rvyl$-space to further alleviate the burden of EBM sampling and learning.

\textbf{$\rvyl$-space for conditional EBM prior.} For our diffusion model, we formulate the marginal EBM prior (Eqn. \ref{marginal-y-ebm}) to a sequence of conditional EBMs prior, where each marginal EBM prior at each diffusion step (i.e., $p_{\omega, \beta_{>0}}(\rvyl_t)$) is constrained by the forward generated $\rvyl_{t+1}$, i.e.,
\begin{eqnarray}\label{cond-y-ebm}
 p_{\omega, \beta_{>0}}(\rvyl_t|\rvyl_{t+1}) \propto p_{\omega, \beta_{>0}}(\rvyl_t)p(\rvyl_{t+1}|\rvyl_{t}) = 
\end{eqnarray}
\begin{equation}
\frac{1}{\mathrm{Z}_{\omega, \beta_{>0}}(\rvyl_{t+1})}\exp \left[F_\omega(T_{\beta_{>0}}(\rvyl_t), t)\right]p_0(\rvyl_t) \cdot p(\rvyl_{t+1}|\rvyl_{t})\nonumber
\end{equation}
where we slightly abuse the notation and use $p(\rvyl_{t+1}|\rvyl_{t})$ for the perturbation kernel as in Eqn. \ref{diff-forward-kernel-y}. The energy function essentially couples all layers of latent variables (i.e., $F_\omega(T_{\beta_{>0}}(\rvyl_t), t)=\sum_{i=1}^L f_{\omega_i}(\rvz_i, t)$), and thus the inter-layer and intra-layer relation of each $\rvz_i$ can be effectively modelled. Each $f_{\omega_i}(\rvz_i, t)$ corresponding to $\rvz_i$ at each layer can also capture the representation of different layers. 

Compared to Eqn. \ref{marginal-y-ebm} that directly models the complex $\rvyl_0$ (generator posterior) with a single marginal EBM, the conditional EBM only models $\rvyl_t$, reversing step by step until $\rvyl_0$. $p(\rvyl_{t+1}|\rvyl_{t})$ from Gaussian noise perturbation kernel can serve to localize $\rvyl_t$ to $\rvyl_{t+1}$, making our conditional EBM less multi-modal and easier to be sampled than the marginal EBM. The reverse trajectory constitutes
\begin{equation}\label{diff-reverse-traj-y}
\begin{aligned}
    p_{\omega, \beta_{>0}}(\rvyl_{0:T}) = \prod_{t=0}^{T-1} p_{\omega, \beta_{>0}}(\rvyl_t|\rvyl_{t+1})p(\rvyl_{T})
\end{aligned}
\end{equation}
where $p(\rvyl_{T}) \sim N(0, \I_{|d|})$ is standard Gaussian as $q(\rvyl_T)$ in forward trajectory.

\textbf{EBM learning.} The proposed method now contains a sequence of parameterized EBMs prior. With forward trajectory Eqn. \ref{diff-forward-traj-y}, we minimize $\KL (q_{\beta}(\rvyl_{0:T}|\rvx) || p_{\omega, \beta_{>0}}(\rvyl_{0:T}))$ for EBM learning. The gradient is estimated as 
\begin{equation}
\label{all-grad-cond-ebm}
\begin{aligned}
    &\nabla_{\omega} \E_{q_{\beta_{>0}}(\rvyl_{0:T}|\rvx)}[\sum_{t=0}^{T-1} F_{\omega}(T_{\beta_{>0}}(\rvyl_t), t) -\\
    &\E_{p_{\omega, \beta_{>0}}(\rvyl_{0:T})}[\sum_{t=0}^{T-1} F_{\omega}(T_{\beta_{>0}}(\rvyl_t),t)]]
\end{aligned}
\end{equation}
which involves sampling from the whole forward and reverse trajectories. To provide more efficient sampling and learning, we follow the strategy used in \cite{ho2020denoising} and utilize random diffusion steps $t$ at each iteration of optimization. Thus, the learning gradient is
\begin{equation}
\label{grad-cond-ebm}
\begin{aligned}
    &\E_{q_{\beta_{>0}}(\rvyl_{t},\rvyl_{t+1}|\rvx)}[\nabla_\omega F_{\omega}(T_{\beta_{>0}}(\rvyl_t), t) - \\
    &\E_{p_{\omega, \beta_{>0}}(\rvyl_t|\rvyl_{t+1})}[\nabla_\omega F_{\omega}(T_{\beta_{>0}}(\rvyl_t),t)]]
\end{aligned}
\end{equation}
where we only need perturbed sample $\rvyl_{t},\rvyl_{t+1}$ from forward trajectory (see Eqn. \ref{diff-forward-traj-y}) and prior samples $\rvyl_t$ from the conditional EBM $p_{\omega, \beta_{>0}}(\rvyl_t|\rvyl_{t+1})$. To obtain prior samples, we perform Langevin dynamics (Eqn. \ref{eq:ebm_langevin}) with the gradient
\begin{eqnarray}\label{eq:ebm_langevin_cond}
   &\nabla_{\rvyl_{t}}\log p_{\omega, \beta_{>0}}(\rvyl_t|\rvyl_{t+1}) = \\
   &\nabla_{\rvyl_{t}} [F_\omega(T_{\beta_{>0}}(\rvyl_t), t)-\frac{||\rvyl_{t}||^2}{2} -\frac{||\alpha_{t+1}\rvyl_{t} - \rvyl_{t+1}||^2}{2\sigma^2_{t+1}}] \nonumber
\end{eqnarray}
Compared to MCMC sampling of marginal EBM (see Eqn. \ref{marginal-y-ebm}) that can be hard to mix between different local modes with noise initial points, MCMC sampling of conditional EBM can start from the given $\rvyl_{t+1}$ and only needs to search for the local modes around $\rvyl_{t+1}$. Specifically, the quadratic term is from the noise-aware term in Eqn. \ref{cond-y-ebm} and constrains the exploration of energy landscape to be localized around $\rvyl_{t+1}$, which is much easier to obtain EBM samples than marginal EBMs. This conditional sampling can be more effective and efficient, which in turn benefits the learning of the proposed EBM prior.

The overall training algorithm and sampling process are shown in Alg. \ref{alg:learning} and Alg. \ref{alg:sampling}.

\subsection{Coupling with symbol vector}
In this section, we present the applicability of the proposed EBM prior. For multi-layer generator models, they are typically learned in an unsupervised scheme and thus are only feasible to generate random samples. However, the controllable and compositional ability to generate desired synthesis nowadays becomes a key requirement for many downstream tasks, yet it would be computationally expensive to \textit{re-train} these models to fit the job. 

The proposed method is flexible in coupling labels or attributes, which in turn empowers controllability and composition ability for the learned multi-layer generator models. Specifically, we can adapt our model with given labels $\rvs$ as
\begin{equation}\label{cond-y-ebm-symbol-coupling}
\begin{aligned}
 &p_{\omega, \beta_{>0}}(\rvyl_t, \rvs|\rvyl_{t+1}) = \frac{1}{\mathrm{Z}_{\omega, \beta_{>0}}(\rvyl_{t+1})} \cdot\\
 &\exp \left[\langle F_\omega(T_{\beta_{>0}}(\rvyl_t), t), \rvs\rangle\right]p_0(\rvyl_t)p(\rvyl_{t+1}|\rvyl_{t})
\end{aligned}
\end{equation}
where the energy function couples both $\rvs$ and $\rvyl_t$, forming an associative memory that allows sampling $\rvyl_t$ with a given $\rvs$. Generating images by such sampled $\rvyl_t$ is known as the controllable generation. \cite{pang2020semi,yu2022latent} adopt similar formulations but only focus on single-layer latent space, while we focus on a multi-layer latent space with a hierarchical structure, allowing coupling $\rvs$ at different layers for specific tasks. We formulate such EBM prior at diffusion step $t=0$ as the signal of $\rvs$ correlates strongly with clean $\rvyl_0$ samples, while noisy $\rvyl_{t>0}$ can be less correlated with $\rvs$. 

The proposed EBM prior is capable of capturing inter-layer and intra-layer relations for different layers of latent variables, thus rendering better controllability and compositionality with effectively learned latent representations. We refer to learning derivation and details in App. \ref{appendix-couple}.

\begin{algorithm}[h]
\begin{algorithmic}[1]
	\REQUIRE ~~\\
	Training images $\rvx$;
	Number of learning iterations $M$; 
 	Hierarchical generator model $\beta$;
        Diffusion steps $T$;
        Langevin steps $k$;.
	\STATE Let $m \leftarrow 0$, initialize EBM parameters $\omega$.  
	\REPEAT 
 	\STATE {\bf Sample $\rvyl_0$:} obtain $\rvzl \sim p_{\beta}(\rvzl|\rvx)$ and $\rvyl_0 = T_{\beta_{>0}}(\rvzl)$
  
        \STATE {\bf Diffusion step $t$:} $t\sim U(0, T-1)$ 
	
	\STATE {\bf Noise sample:} sample $\rvyl_t^*$, $\rvyl_{t+1}^*$ from Eqn. \ref{diff-forward-traj-y}.
 
	\STATE {\bf Prior sample:} sample $\rvyl_t^k$ by Eqn. \ref{eq:ebm_langevin} and Eqn. \ref{eq:ebm_langevin_cond}  with $\rvyl_{t+1}^*$ being the initialization 

        \STATE {\bf Learn $\omega$:} Update $\omega$ with $\rvyl_t^*$ and $\rvyl_t^k$ using Eqn. \ref{grad-cond-ebm}
	\STATE Let $m \leftarrow m+1$.
	\UNTIL $m = M$
\end{algorithmic}
\caption{Learning EBM parameter $\omega$}
\label{alg:learning}
\end{algorithm}

\begin{algorithm}[h]
\begin{algorithmic}[1]
	\REQUIRE ~~\\
        Diffusion steps $T$;
 	Hierarchical generator model $\beta$;
        EBM prior $\omega$.
	\STATE Let $t \leftarrow T-1$ and $\rvyl_{T} \sim N(0, \I_{|d|})$.  
	\REPEAT 
 	\STATE {\bf Sample $\rvyl_t$:} sample $\rvyl_t \sim p_{\omega, \beta_{>0}}(\rvyl_t|\rvyl_{t+1})$ using Eqn. \ref{eq:ebm_langevin} and Eqn. \ref{eq:ebm_langevin_cond} 
	\STATE Let $t \leftarrow t-1$.
	\UNTIL $t = -1$
        \STATE {\bf Sample $\rvzl$:} obtain $\rvzl = T^{-1}_{\beta_{>0}}(\rvyl_0)$.  
        \STATE {\bf Generate $\rvx$:} generate $\rvx \sim p_{\beta_0}(\rvx|\rvzl)$ with obtained $\rvzl$. 
\end{algorithmic}
\caption{Sampling and Image Synthesis}
\label{alg:sampling}
\end{algorithm}

\section{Related Work}
\textbf{Energy-based model.} The EBM is expressive in representing the data uncertainty. Most existing works learn EBM on data space by maximizing EBM likelihood, which involves challenging MCMC sampling for the EBM. To tackle the challenge, \cite{du2019implicit,du2020improved} propose the use of a replay buffer, while \cite{cui2023nips,han2020joint,xie2021learning,xie2018cooperative} consider learning a complementary generator model to jump-start MCMC sampling. \cite{xiao2022adaptive,pang2020learning,pang2020semi} focus on learning EBM prior on low-dimensional latent space to mitigate the burden of EBM sampling, but these works only deal with a single-layer latent space. We consider learning EBM prior for multi-layer latent variables, which enables hierarchical representation learning and better sample generation. 

\textbf{Hierarchical generative model.} For hierarchical generative models \cite{vahdat2020nvae,maaloe2019biva,NIPS2016_6ae07dcb}, they typically assume a Gaussian prior model, which can be less informative, resulting in the prior hole problem and poor generation quality. Our work aims to learn effective EBM prior for hierarchical generative models, which is related to joint EBM prior (see Sec. \ref{sec-bg-jebm}). These works intend to learn a single (marginal) EBM on $\rvzl$-space, while we study learning a sequence of conditional EBMs on $\rvyl$-space. NCP-VAE \cite{aneja2021contrastive} discuss a \textit{autoregressive}-style model, which can be \textit{intractable} for MLE learning, as the normalization constant includes the top layer $\rvz_{i+1}$ and needs an additional inner loop for sampling. Our model leverages diffusion probabilistic models such that we can directly draw $\rvyl_{t+1}$ through the forward trajectory. Compared to \cite{Cui_2023_CVPR} (Eqn. \ref{joint-ebm}), our EBM prior only models $\rvyl_t$ conditioned on the perturbed sample $\rvyl_{t+1}$ generated by $\rvyl_{t+1} = \alpha_{t+1}\rvyl_t + \epsilon_{t+1}\sigma_{t+1}$, which can be much easier than directly recovering from $\rvyl_{T}$ to $\rvyl_0$ (if consider equivalent form Eqn. \ref{marginal-y-ebm}).

\textbf{EBM diffusion model.} Other diffusion EBMs \cite{gao2020learning,yu2022latent,yu2023diffusionAmortized} motivate EBM learning with a diffusion scheme. They build energy-based \textit{recovery} model on data space and single-layer latent space, respectively. In this work, we focus on multi-layer generator model with a top-down hierarchical structure, which is shown to be capable of learning meaningful hierarchical representations \cite{child2020very,zhao2017learning,havtorn2021hierarchical}. It is typically challenging for the diffusion scheme to maintain such hierarchical structures as the goal of the forward process is to destroy the data pattern, which includes the inter-layer relation (conditional dependency) among multi-layer latent variables. To preserve the hierarchical structure of latent variables, we conduct the forward and reverse processes on $\rvyl$-space, which also leads to more effective learning and sampling than multi-scale $\rvx$-space \cite{gao2020learning} and $\rvz$-space \cite{yu2022latent}.

\begin{figure*}[t]
\centering
\includegraphics[width=0.675\textwidth]{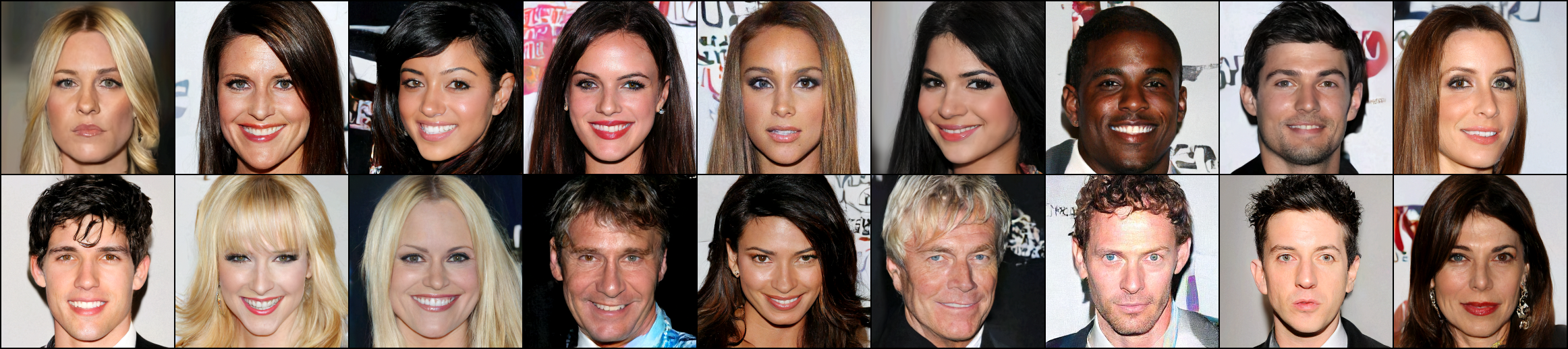}
        \includegraphics[width=0.315\columnwidth]{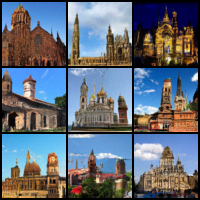}
        \includegraphics[width=0.315\columnwidth]{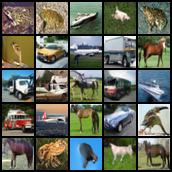}
\caption{Image synthesis on CelebA-HQ-256 (left), LSUN-Church-64 (center) and CIFAR-10 (right).}
\label{Fig.celeba256-syn}
\end{figure*}


\begin{minipage}[t]{0.99\columnwidth}
    \centering
    \begin{minipage}{0.99\columnwidth}
        \centering
        \captionof{table}{IS$(\uparrow)$ and FID$(\downarrow)$ on CIFAR-10. Model$^*$ indicates our backbone model.}
        \label{table.cifar-backbone-fid}
        \resizebox{0.9\columnwidth}{!}{
          \begin{tabular}{lcc}
        \toprule
            Method  & IS & FID\\
        \toprule
            \textbf{Ours} ($T=3$) & 9.03 & 8.93 \\
            Joint-EBM \cite{Cui_2023_CVPR} & 8.99 & 11.34 \\
            DRL-EBM ($T=6$) \cite{gao2020learning} & 8.40 & 9.58 \\
            NCP-VAE \cite{aneja2021contrastive} & - & 24.08\\
        \midrule
        \textbf{Hierarchical Generative Models w Gaussian Prior} & &\\
            NVAE$^*$ \cite{vahdat2020nvae} & 5.30 & 37.73 \\
            NVAE$^*$-Recon & - & 0.68 \\
            HVAE \cite{NIPS2016_6ae07dcb} & - & 81.44 \\
            BIVA \cite{maaloe2019biva} & - & 66.37 \\
        \midrule
        \textbf{Energy-based Models} & &\\
            Architectural-EBM \cite{Cui_2023_ICCV} & - & 63.42 \\
            Dual-MCMC \cite{cui2023nips} & 8.55 & 9.26 \\
            Adaptive-CE \cite{xiao2022adaptive} & - & 65.01 \\
            VAEBM \cite{xiao2020vaebm} & 8.43 & 12.19 \\
            Hat EBM \cite{hill2022hatebm} & - & 19.15\\
            ImprovedCD \cite{du2020improved} & 7.85 &25.1 \\
            Divergence Triangle \cite{Han_2020_CVPR} & - &30.10 \\
            Adv-EBM \cite{yin2020advebm} & 9.10 & 13.21 \\
        \toprule
            \textbf{GANs}+\textbf{Score}+\textbf{Diffusion Models} && \\
            StyleGANv2 w/o ADA \cite{karras2020training} & 8.99 & 9.9 \\
            Diffusion-Amortized \cite{yu2023diffusionAmortized} & - & 57.72\\
            NCSN \cite{song2019generative} & 8.87 & 25.32 \\
            LSGM \cite{vahdat2021score} & - & 2.10\\
            DDPM ($T=1000$) \cite{ho2020denoising} & 9.46 & 3.17\\
        \toprule
          \end{tabular}
        }
    \end{minipage}
    \begin{minipage}{0.99\columnwidth}
        \centering
        \captionof{table}{FID on CelebA-HQ-256 and LSUN-Church-64.}
        \label{table.celeba256-backbone-fid}
        \resizebox{0.99\columnwidth}{!}{
          \begin{tabular}{l c c}
        \toprule
            Method & CelebA-HQ-256 & LSUN-Church-64\\
        \toprule
            \textbf{Ours} ($T=3$) & 8.78 & 7.34\\
            Joint-EBM \cite{Cui_2023_CVPR} & 9.89 & 8.38\\
            DRL-EBM ($T=6$) \cite{gao2020learning} & - & 7.04 \\
            NCP-VAE \cite{aneja2021contrastive} & 24.79 & -\\
        \toprule
            NVAE$^*$ \cite{vahdat2020nvae} & 30.25 & 38.13\\
            NVAE$^*$-Recon & 1.64 & 2.45 \\
        \cmidrule(lr){0-0}
            Adv-EBM \cite{yin2020advebm} & 17.31 & 10.84\\
            GLOW \cite{kingma2018glow} & 68.93 & 59.35\\
            PGGAN \cite{karras2017progressive} & 8.03 & 6.42\\
        \bottomrule
          \end{tabular}
        }
    \end{minipage}
\end{minipage}

\section{Experiment}
\subsection{Image Synthesis}\label{exp-img-syn}
First, we examine the sample quality of our model. The proposed model is learned as a reverse approximation model in a diffusion probabilistic scheme, which allows sampling reverse steps $\rvyl_t$ to reach $\rvyl_0$. With sampled $\rvyl_0$, we obtain $\rvzl_0$ through the transformation function and then generate images $\rvx$ through the generator model (see Alg. \ref{alg:sampling}). \blfootnote{Our project page is available at \url{https://jcui1224.github.io/diffusion-hierarchical-ebm-proj/}.}

We assess our model on the standard benchmark CIFAR-10 and the challenging high-resolution CelebA-HQ-256 and large-scale LSUN-Church-64. We compare with our direct baseline model Joint-EBM \cite{Cui_2023_CVPR}, NCP-VAE \cite{aneja2021contrastive} that learn signal (marginal) EBM prior for hierarchical generative models, and Diffusion Recovery (DRL) EBM \cite{gao2020learning} which learn EBM with diffusion probabilistic scheme on data space, as well as hierarchical generative model with the Gaussian prior and other powerful advanced generative models. 

We recruit Fr$\acute{e}$Chet Inception Distance (FID) and Inception Score (IS) metrics to evaluate the quality of image synthesis. We report our results in Tab. \ref{table.cifar-backbone-fid} and Tab. \ref{table.celeba256-backbone-fid} as well as the FID score of the reconstructed images. It can be observed that our hierarchical EBM prior shows superior performance compared to our baseline models and can even be competitive with those powerful GANs and diffusion-based methods. For a fair comparison, we adopt the NVAE model \cite{vahdat2020nvae} as the backbone generator model as our direct baseline Joint-EBM and NCP-VAE. More quantitative and qualitative results can be found in the ablation studies and App. \ref{sec-app-exp}.

\subsection{Hierarchical Representation}
Different from other generator models, the hierarchical generator model has an appealing structure, in which the latent variables at the top layers tend to learn high-level semantic representations, while low-level details representation can be learned by latent variables at lower layers. We examine our model in learning such hierarchical representations. 

\textbf{Hierarchical sampling.} First, we demonstrate our model by performing \textit{hierarchical sampling}, which generates variations of image synthesis that can visualize the different levels of data representation learned by different layers of latent variables. In particular, for multi-layer latent variables, we only generate random samples for some layers of latent variables while fixing other layers; hence, the corresponding variation of features is captured by the latent variables randomly sampled. We show visualization results on CelebA-HQ-256 in Fig. \ref{Fig.hs-celeba256}. It can be observed that by randomly sampling for the top layers of latent variables, the general structure (e.g., genders and face identities) would be changed, while for lower layers of latent variables, low-level features (e.g., hair color, skin color) can vary correspondingly. We note that this is a challenging task where some minority of features are entangled across layers, but the majority of features can be successfully captured. This showcases the capability of our model in learning hierarchical representations.

\begin{minipage}{\columnwidth}
    \centering
    \includegraphics[width=0.99\columnwidth]{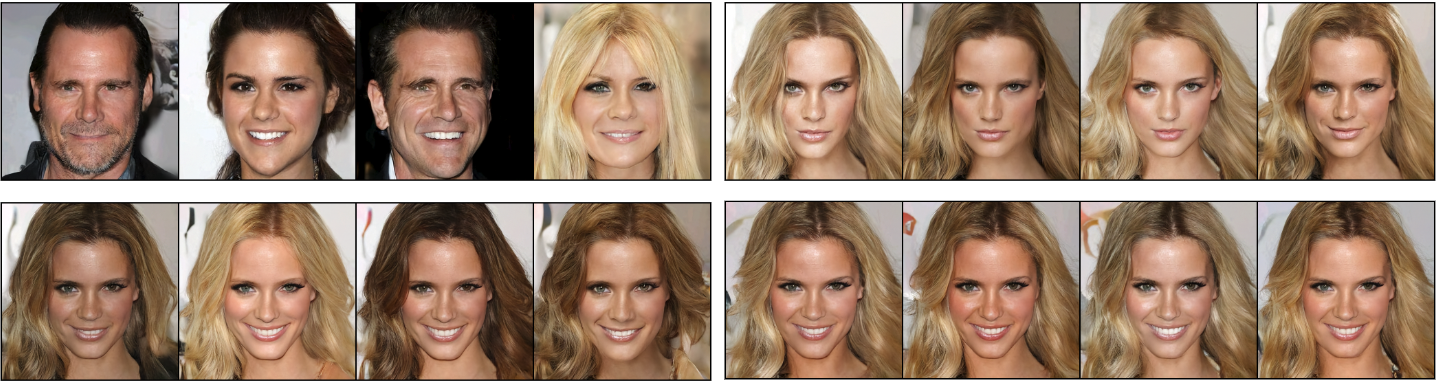}
    \captionof{figure}{Hierarchical sampling. Visualization of representations learned by latent variables from the top to bottom layers, arranged as top-left, top-right, bottom-left and bottom-right.}
    \label{Fig.hs-celeba256}
\end{minipage}

\textbf{Out-of-distribution detection.} Then, we evaluate our model in out-of-distribution (OOD) detection task to further demonstrate learned hierarchical representations. Typically, EBMs can be applied to the OOD task by computing the energy score as the decision function \cite{hill2022learning,cui2023nips}. For latent space EBMs, the inferred latent sample (i.e., $\rvyl_0 \sim q_\beta(\rvyl_0|\rvx)$ for our case) from in-distribution (ID) data is usually assigned with a lower energy than from the OOD data. In this work, we compute the energy score of inferred latent samples at the top layers as the decision function for using high-level semantic representations learned at the top layers of latent variables to distinguish the OOD and the ID data \cite{havtorn2021hierarchical}. 

To better leverage the diffusion probabilistic scheme, we propose conducting EBM sampling based on perturbed inference samples at the top layers, together with prior samples at the bottom layers. Specifically, with testing images $\rvx$ and layer index $k$, we first obtain $\rvyl_1^{>k}\sim q_{\beta_{>0}}(\rvyl_{0:T}|\rvx)$ (see Eqn. \ref{diff-forward-traj-y}) and $\rvyl_1^{\leq k}\sim p_{\omega,\beta_{>0}}(\rvyl_{0:T})$ (see Eqn. \ref{diff-reverse-traj-y}) where we choose diffusion step $t=1$ to ensure that only minor noise is added. Then, we perform reverse sampling conditioned on $\rvyl_1^{>k}$ and $\rvyl_1^{\leq k}$ for jointly sampling final $\rvyl_0$, i.e., $\rvyl_0 \sim p_{\omega,\beta_{>0}}(\rvyl_{0}|[\rvyl_1^{>k}, \rvyl_1^{\leq k}])$ where $[,]$ is the operation of concatenation. If $\rvyl_1^{>k}$ is from ID data, then $\rvyl_0^{>k}$ should render lower energy scores as both the $\rvyl_1^{>k}$ and $\rvyl_1^{\leq k}$ are from similar local modes of the learned energy landscape. If $\rvyl_1^{>k}$ is from OOD data, then $\rvyl_1^{>k}$ and $\rvyl_1^{\leq k}$ can be in different modes, making the final reverse sampling difficult to traverse the energy landscape and thus rendering higher energy score of sampled $\rvyl_0^{>k}$. We follow the standard protocol and evaluate by AUROC score for our EBM prior trained on CIFAR-10 with SVHN dataset being the OOD dataset. The result is reported in Fig. \ref{Fig.ood} where the performance of our model indeed improves as the layer of $k$ increases, which agrees with the observation in \cite{havtorn2021hierarchical} that our model can capture hierarchical representations at different layers. Compared to using inferred latent samples (inference scheme in Fig. \ref{Fig.ood}), the proposed diffusion-based method (diffusion scheme in Fig. \ref{Fig.ood}) can render better performance by conducting additional MCMC sampling of the learned EBM.

\begin{minipage}{\columnwidth}
    \begin{minipage}{0.95\columnwidth}
        \centering
        \includegraphics[width=0.99\columnwidth]{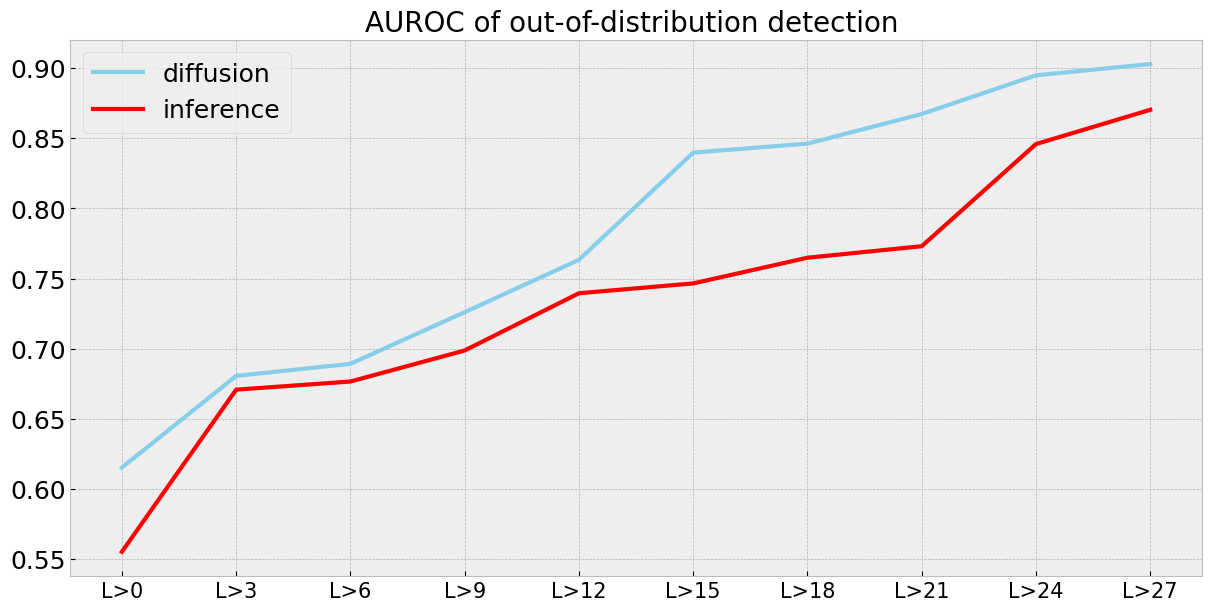}
    \end{minipage}
    \begin{minipage}{0.95\columnwidth}
        \centering
        \includegraphics[width=0.99\columnwidth]{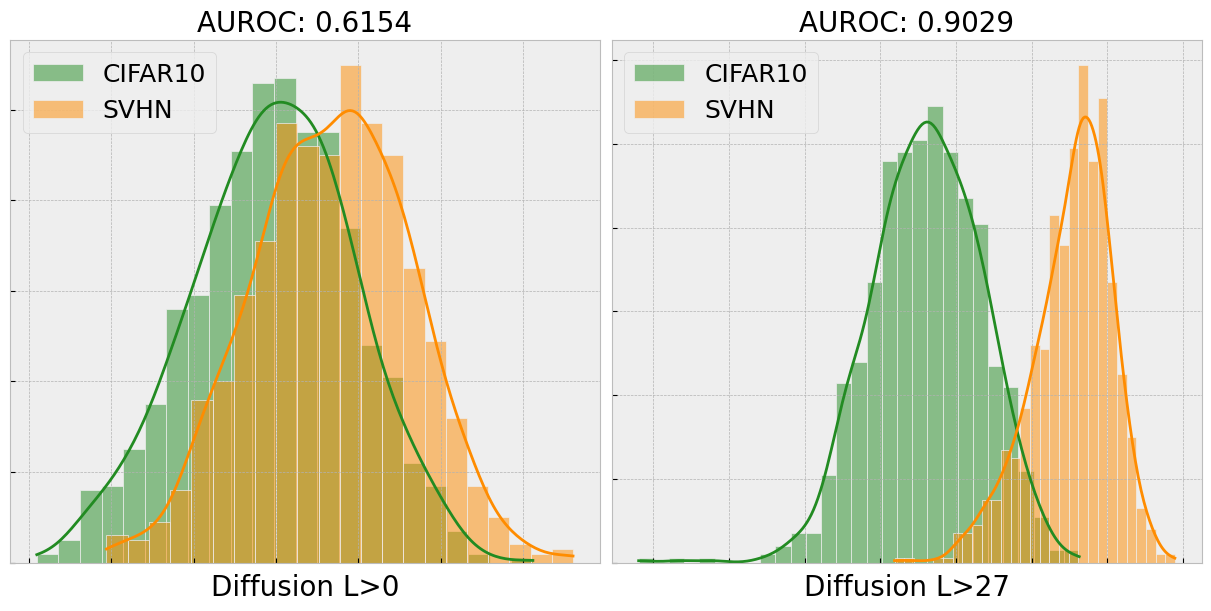}
        \captionof{figure}{AUROC results for energy scores of different layers (denoted as $L>k$ for using top layers above $k$-th layer). \textbf{Top} figure visualizes the comparison between the diffusion scheme ($\rvyl_0$ sampled from EBM) and the inference scheme ($\rvyl_0$ inferred from inference model) in different layers. \textbf{Bottom} figure is the histogram of energy scores using all layers $L>0$ and top layers $L>27$. Total number of layers is 30.}
        \label{Fig.ood}
    \end{minipage}
\end{minipage}

\begin{figure*}[t]
\centering
\includegraphics[width=0.99\textwidth]{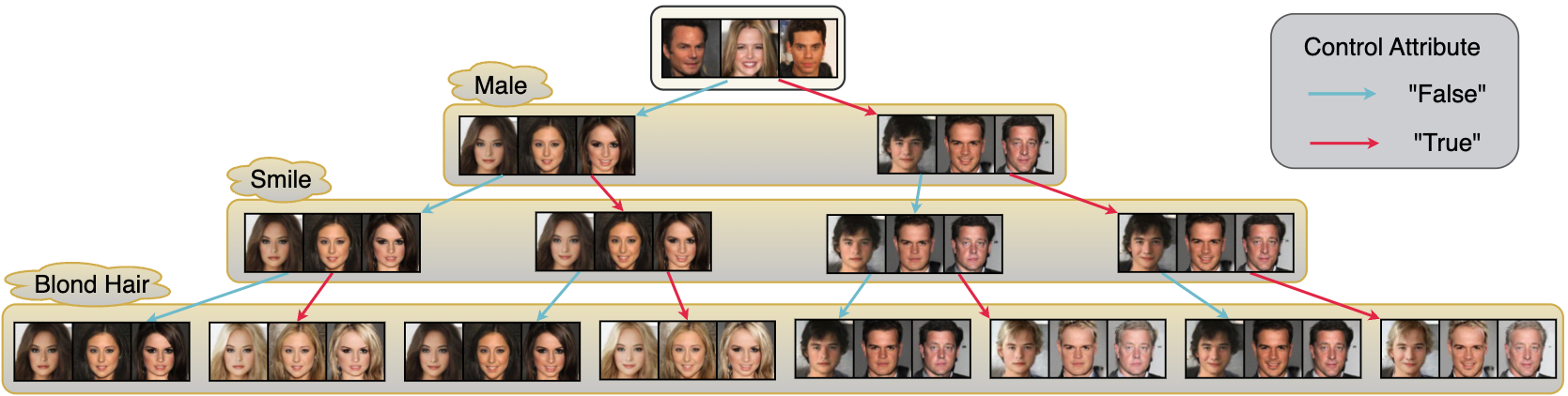}
\caption{Fine-tuned image synthesis with multiple attributes on CelebA-64.}
\label{Fig.fine-tuned}
\end{figure*}

\subsection{Controllable Synthesis}
For hierarchical generator models that are typically learned without labels (unsupervised learning), they can only generate random synthesis. Our diffusion EBM prior, as a flexible complementary model, can be coupled with labels to make hierarchical generative models more applicable in downstream tasks, such as controllable and compositional generation. In practice, we fix the hierarchical generative model and train our adapted model (Eqn. \ref{cond-y-ebm-symbol-coupling}) with given $\rvs$ (supervised learning), and then we assess if our model can render controllability and compositionality with the captured hierarchical representations. We refer to App. \ref{appendix-couple} for details of training and sampling.

\textbf{Categorical labels.} First, we evaluate our model with categorical label information, e.g., $\rvs$ represents labels of objects in CIFAR-10. To better examine our model, we only consider coupling $\rvs$ at the top layers by which the semantic representations are captured. With the learned model, we conduct controllable generation by specifying target labels $\rvs$ and sampling corresponding $\rvyl$  for generating the images. We show the results in Fig. \ref{Fig.control-syn}, and it can be seen that our model can capture data representations and thus can generate synthesis with specific categories.

\begin{minipage}{\columnwidth}
    \centering
    \includegraphics[width=0.99\columnwidth]{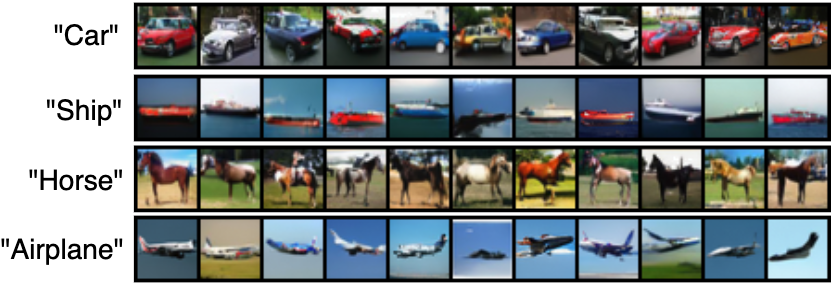}
    \captionof{figure}{Controllable synthesis on CIFAR-10.}
    \label{Fig.control-syn}
\end{minipage}

%

\textbf{Multiple attributes.} Then, we showcase our model by utilizing multiple data attributes for the challenging \textit{fine-tuned} image synthesis. In particular, for example of CelebA-64 dataset, we can have multiple attributes $\rvs_{1:N}$ for different levels of data features. To better suit the hierarchical structure, we choose to couple high-level attributes, such as gender information, at the top layers and gradually couple lower-level attributes, such as face and hair features, at the lower layers. In Fig. \ref{Fig.fine-tuned}, we first specify gender attributes (e.g., "Male") and sample $\rvyl$ at the corresponding top layers for generating gender-specified images. Then, we fix these $\rvyl$ of top layers and specify lower-level attributes (e.g., "Smile" and "Blond Hair") for sampling $\rvyl$ at lower layers. We observe our EBM prior successfully renders the desired compositional synthesis by gradually adding specific features to the sampled images without changing the majority of previously fixed features. This suggests learned hierarchical representation of our EBM prior.

\subsection{Langevin Transition for Energy Landscape}
\begin{figure*}
    \centering
    \includegraphics[width=0.99\textwidth]{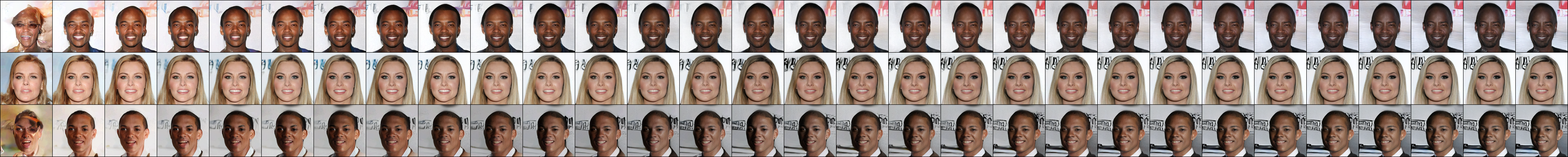}
    \caption{Long-run Langevin transition on CelebA-HQ-256. Visualization for each 30 steps.}
    \label{Fig.add-langevin-trans-long}
\end{figure*}
We examine the energy landscape of our learned EBM prior by visualizing the Langevin transition. Our conditional EBM prior should render a smooth energy landscape such that Langevin dynamics can effectively explore with a smooth Langevin transition. We visualize the corresponding image synthesis of the Langevin trajectory of each diffusion step, i.e., $t=T-1, \dots, 1, 0$. We show in Fig. \ref{Fig.add-langevin-trans} for each step of the short-run Langevin dynamics (e.g., 30 Langevin steps) and in Fig. \ref{Fig.add-langevin-trans-long} for a challenging long-run setting (e.g., 300 Langevin steps for each diffusion step). We observe in Fig. \ref{Fig.add-langevin-trans} that the quality of image synthesis becomes better as the Langevin progresses, with large improvement at diffusion step $t=T-1$ and minor improvement at diffusion step $t=0$; while, in Fig. \ref{Fig.add-langevin-trans-long}, we do not see an \textit{oversaturated problem} of EBM learning as observed in \cite{nijkamp2020anatomy}. We conduct such experiments to demonstrate a smooth energy landscape learned for our EBMs prior. 

\begin{minipage}{\columnwidth}
    \centering
    \includegraphics[width=0.99\columnwidth]{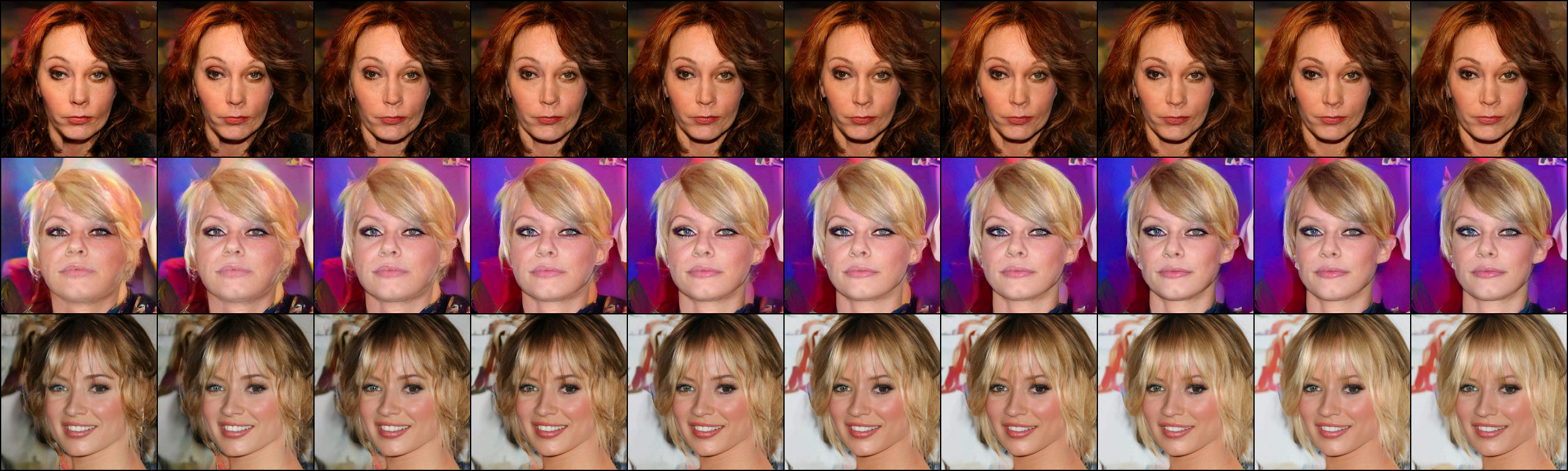}
    \includegraphics[width=0.99\columnwidth]{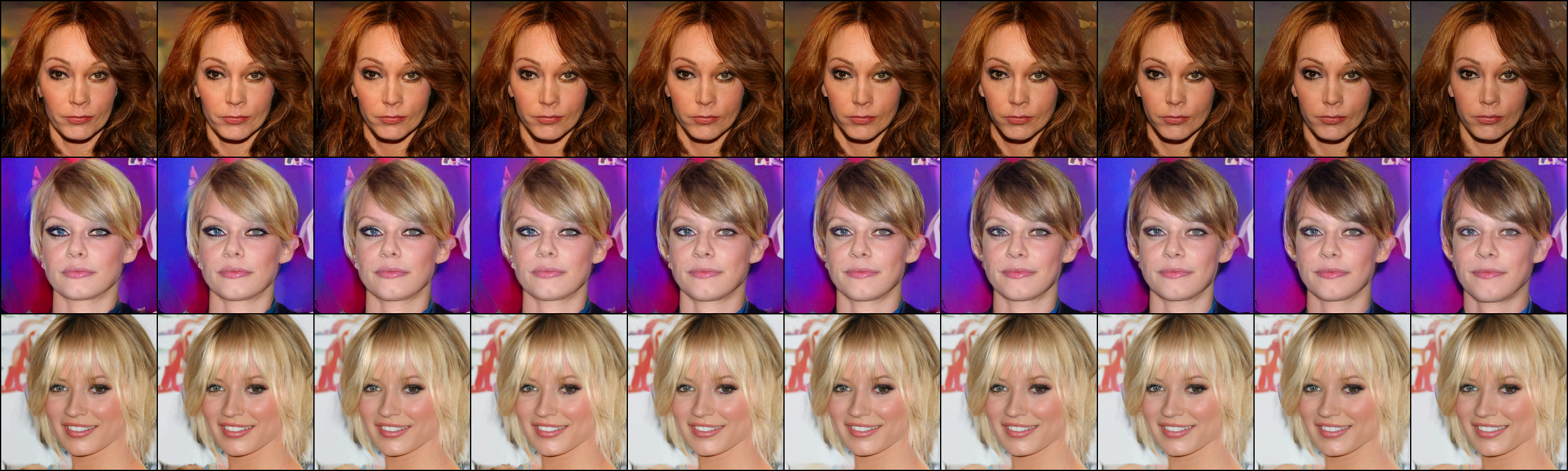}
    \includegraphics[width=0.99\columnwidth]{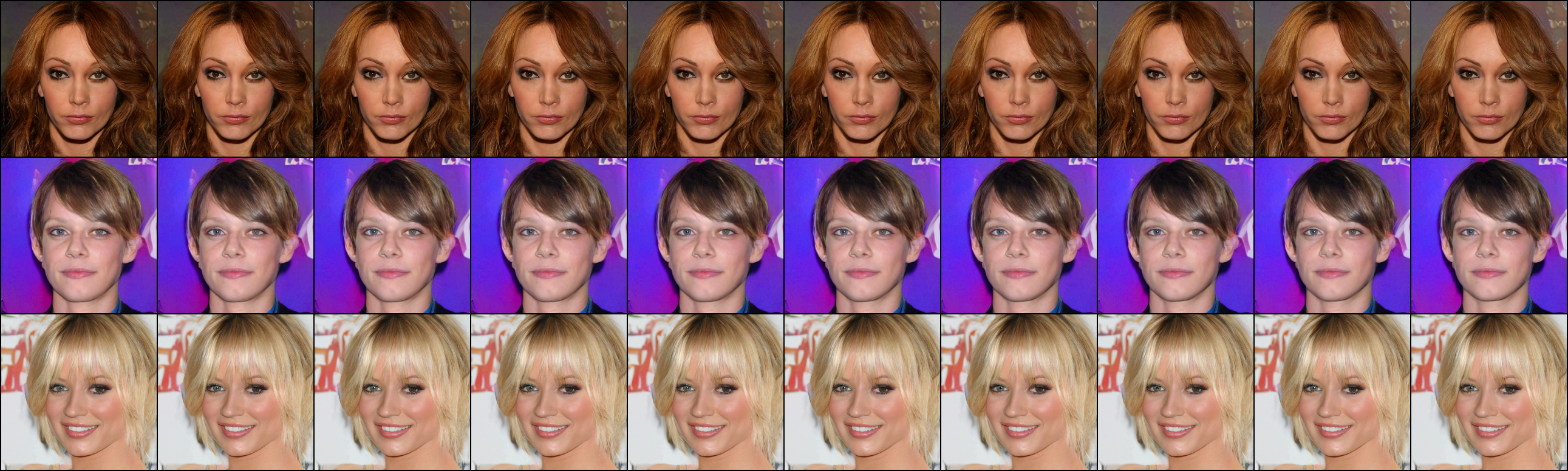}
    \captionof{figure}{Short-run Langevin transition on CelebA-HQ-256. The top rows of figures show the transition for the diffusion step at $t=T-1$, and the bottom rows show the transition for the diffusion step at $t=0$.}
    \label{Fig.add-langevin-trans}
\end{minipage}

\subsection{Ablation Studies}
\textbf{Diffusion step $T$.} First, we train our diffusion-based EBM prior with more diffusion steps, e.g., $T=6$. By doing so, our model should render better performance with easier EBM sampling by matching less perturbed samples at each step. We report the FID score and sampling time in Tab.\ref{table.LD-steps} where the synthesis quality indeed improves with more diffusion steps but also requires more sampling time. We thus report $T=3$ as our result.

\begin{minipage}{\columnwidth}
    \centering
    \captionof{table}{Langevin steps $K$ and diffusion steps $T$.}
    \label{table.LD-steps}
    \resizebox{0.99\columnwidth}{!}{
      \begin{tabular}{ccc|c|cc}
    \toprule
         & $K=30$ & $K=100$ & $K=50$, $T=3$ & $T=6$ \\
    \toprule
         FID & 9.98 & 8.13 & \textbf{8.93} & 8.13\\
         Time (seconds) & 75.17 & 166.34 & 94.23 & 193.45\\
    \toprule
      \end{tabular}
    }
\end{minipage}

\textbf{Langevin step $K$.} By using more Langevin steps, we should explore the energy landscape better and obtain more effective EBM samples for learning. The learned EBMs can thus generate high-quality samples for image synthesis. We show our results in Tab. \ref{table.LD-steps} where using 50 steps (denoted as $K=50$) delivers better synthesis than using 30 steps, while using 100 steps only shows a minor improvement but costs much more training and sampling time.

\begin{minipage}{\columnwidth}
    \centering
    \includegraphics[width=0.95\columnwidth]{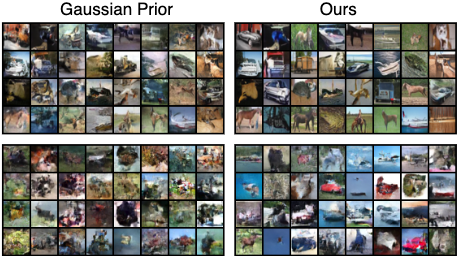}
    \captionof{figure}{Images synthesis on CIFAR-10 with different backbones. Top-row images are generated with the NVAE (30 layers) backbone, and bottom-row images are generated with the HVAE (3 layers) backbone.}
    \label{Fig.hvae}
\end{minipage}

\textbf{Other hierarchical generator models.} In addition to the NVAE backbone model, we also train a simple backbone hierarchical VAE with 3-layer latent variables. We visualize images generated by the Gaussian prior and our model in Fig. \ref{Fig.hvae}, where our model still improves the quality of the generation in a large way (from FID 81.44 to 35.13), suggesting the effectiveness of the proposed method.

\section{Limitation}
In this work, the proposed method still renders inferior performance compared to state-of-the-art models (e.g., modern diffusion probabilistic models \cite{vahdat2021score}). In addition, sampling from our EBM prior requires an iterative Langevin dynamics sampler, which can be further improved or even bypassed; we will consider it in our future studies.

\section{Conclusion}
We propose learning EBM prior for hierarchical generative model with a diffusion probabilistic scheme, which features more tractable conditional likelihood learning and more effective EBM sampling. We employ a uni-scale $\rvyl$-space to maintain the hierarchical structure and further mitigate the burden of MCMC sampling. Such learned EBM prior can generate high-quality samples for image synthesis and can capture hierarchical representations for downstream tasks. 

\section*{Impact Statement}
This paper presents a generative probabilistic framework whose goal is to advance the field of Machine Learning and may share the limitations and negative impact as other advanced generative models. There are many potential societal consequences of our work, none of which we feel must be specifically highlighted here.

\nocite{langley00}

\bibliography{example_paper}
\bibliographystyle{icml2024}

\newpage
\appendix
\onecolumn
\section{Theorectical Derivation}
\subsection{Between $\rvyl$-space and $\rvzl$-space}\label{appendix-derivation}
For conditional Gaussian distribution $p_{\beta_i}(\rvz_i|\rvz_{i+1})$ (see Eqn. \ref{gaussian-multi-joint}), we can have an invertible transformation function $T_{\beta_{>0}}$. For example of 2-layer latent variables, it is defined specifically as
\begin{eqnarray}
\label{t_1}
\rvz_2 = T_{\beta_{>0}}^{\rvz_2}(\rvy_{2}) = \rvy_2 \;\;\text{and}\;\; \rvz_1 = T_{\beta_{>0}}^{\rvz_1}(\rvy_1, \rvy_2) = \mu_{\beta_1}(\rvz_2) + \sigma_{\beta_1}(\rvz_{2}) \cdot\rvy_1
\end{eqnarray}
where $\rvy_1$ and $\rvy_2$ are distributed as independent Gaussian noise, i.e., $(\rvy_1,\rvy_2) \sim p_0(\rvy_1, \rvy_2)$ and $p_0(\rvy_1, \rvy_2) = p_0(\rvy_1)p_0(\rvy_2)$ with each $p_0(\rvy_i) \sim \N(0, I_{d})$. By change-of-variable rule, we can have 
\begin{eqnarray}
\label{t_2}
p_{\beta_{>0}}(\rvz_1, \rvz_2) = p_0(\rvy_1, \rvy_2) | \mathrm{det} (J_{T_{\beta_{>0}}^{-1}})| \;\;\text{and}\;\; p_0(\rvy_1, \rvy_2) = p_{\beta_{>0}}(\rvz_1, \rvz_2) | \mathrm{det} (J_{T_{\beta_{>0}}})|
\end{eqnarray}
where $J_{T_{\beta_{>0}}}$ is the Jacobian of $T_{\beta_{>0}}$. 

For joint EBM prior on $\rvzl$-space $p_{\omega, \beta_{>0}}(\rvzl)$ (see Eqn. \ref{joint-ebm}), we can apply the change-of-variable rule and Eqn. \ref{t_2} as
\begin{equation}
\begin{aligned}
\label{COV}
&p_{\omega, \beta_{>0}}(\rvyl) = p_{\omega, \beta_{>0}}(\rvzl) | \mathrm{det} (J_{T_{\beta_{>0}}})|\\
&= \frac{1}{\mathrm{Z}_{\omega, \beta_{>0}}}\exp [F_{\omega}(T_{\beta_{>0}}(\rvyl))]p_{\beta_{>0}}(\rvzl)| \mathrm{det} (J_{T_{\beta_{>0}}})|\\
&= \frac{1}{\mathrm{Z}_{\omega, \beta_{>0}}}\exp [F_{\omega}(T_{\beta_{>0}}(\rvyl))]p_0(\rvyl)
\end{aligned}
\end{equation}
which is the Eqn. \ref{marginal-y-ebm}. With such marginal EBM prior on $\rvyl$-space, we construct our conditional EBM prior as shown in Eqn. \ref{cond-y-ebm}.

\subsection{Coupling with symbol vector}\label{appendix-couple}
Energy-based model can be flexible to couple with symbol vector \cite{pang2020semi,yu2022latent,nie2021controllable,grathwohl2019JEM}. For Eqn. \ref{cond-y-ebm-symbol-coupling}, we adapt our model to couple with symbol vector $\rvs$. The marginal version of Eqn. \ref{cond-y-ebm-symbol-coupling} is given as
\begin{eqnarray}
\label{t_3}
 \hat{p}_{\omega, \beta_{>0}}(\rvyl_t|\rvyl_{t+1}) = \frac{1}{\mathrm{Z}_{\omega, \beta_{>0}}(\rvyl_{t+1})}
 \exp \left[\hat{F}_\omega(T_{\beta_{>0}}(\rvyl_t), t)\right]p_0(\rvyl_t)p(\rvyl_{t+1}|\rvyl_{t})
\end{eqnarray}
where $\hat{F}_\omega(T_{\beta_{>0}}(\rvyl_t), t) = \log \sum_\rvs \exp (\langle F_\omega(T_{\beta_{>0}}(\rvyl_t), t), \rvs \rangle)$. This forms a softmax classifier, i.e.,
\begin{eqnarray}
\label{t_4}
 p_{\omega, \beta_{>0}}(\rvs|\rvyl_t, \rvyl_{t+1}) = \frac{p_{\omega, \beta_{>0}}(\rvyl_t, \rvs|\rvyl_{t+1})}{\hat{p}_{\omega, \beta_{>0}}(\rvyl_t|\rvyl_{t+1})}=\frac{\exp (\langle F_\omega(T_{\beta_{>0}}(\rvyl_t), t), \rvs \rangle)}{\sum_\rvs \exp (\langle F_\omega(T_{\beta_{>0}}(\rvyl_t), t), \rvs \rangle)}
\end{eqnarray}
where energy function $F_\omega(T_{\beta_{>0}}(\rvyl_t), t)$ outputs the logit score of categories. Recall that $F_\omega(T_{\beta_{>0}}(\rvyl_t), t) = \sum_{i=1}^L f_{\omega_i}(\rvz_i, t)$, we therefore can couple $\rvs$ at different layers and let the energy score $f_{\omega_i}(\rvz_i, t)$ at $i$-th layer serve as the softmax classifier for $\rvs$. 

Recall that we only couple $\rvs$ at $t=0$, learning such model can be achieved by maximizing the likelihood of $\log p_{\omega, \beta_{>0}}(\rvyl_{0:T}, \rvs)$. Specifically, we have
\begin{eqnarray}
\label{t_5}
    \log p_{\omega, \beta_{>0}}(\rvyl_{0:T}, \rvs) &=& \underbrace{\log \hat{p}_{\omega, \beta_{>0}}(\rvyl_0|\rvyl_{1}) + \log p_{\omega, \beta_{>0}}(\rvs|\rvyl_0, \rvyl_{1})}_{\text{with symbol vector}} \\
\label{t_6}
    &+& \underbrace{\sum_{t=1}^{T-1} \log p_{\omega, \beta_{>0}}(\rvyl_t|\rvyl_{t+1})}_{\text{without symbol vector}}
\end{eqnarray}
where Eqn. \ref{t_6} computes the gradient similar as Eqn. \ref{grad-cond-ebm}, while for Eqn. \ref{t_5}, it is learned with an extra term computing the gradient for the softmax classifier $p_{\omega, \beta_{>0}}(\rvs|\rvyl_0, \rvyl_{1})$, i.e., optimizing using standard cross-entropy.

For sampling $\rvyl_0$ with specified $\rvs$, we first obtain $\rvyl_1$ by reversing step by step from $\rvyl_T$ via Langevin dynamics (see Alg. \ref{alg:sampling}). Then, we perform Langevin dynamics (see Eqn. \ref{eq:ebm_langevin}) to sample from $p_{\omega, \beta_{>0}}(\rvyl_0, \rvs|\rvyl_{1})$. The gradient is then computed as $\nabla_{\rvyl_0}\log p_{\omega, \beta_{>0}}(\rvyl_0, \rvs|\rvyl_{1}) = \nabla_{\rvyl_0}\left[\log \hat{p}_{\omega, \beta_{>0}}(\rvyl_0|\rvyl_{1}) + \log \hat{p}_{\omega, \beta_{>0}}(\rvs|\rvyl_0,\rvyl_{1})\right]$

\section{Additional Result}\label{sec-app-exp}
\subsection{Additional Result}
\begin{minipage}{\columnwidth}
    \centering
    \includegraphics[width=0.99\columnwidth]{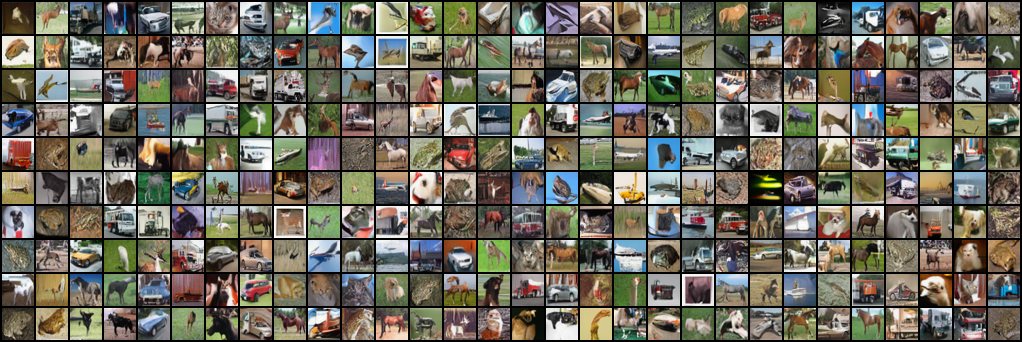}
    \includegraphics[width=0.99\columnwidth]{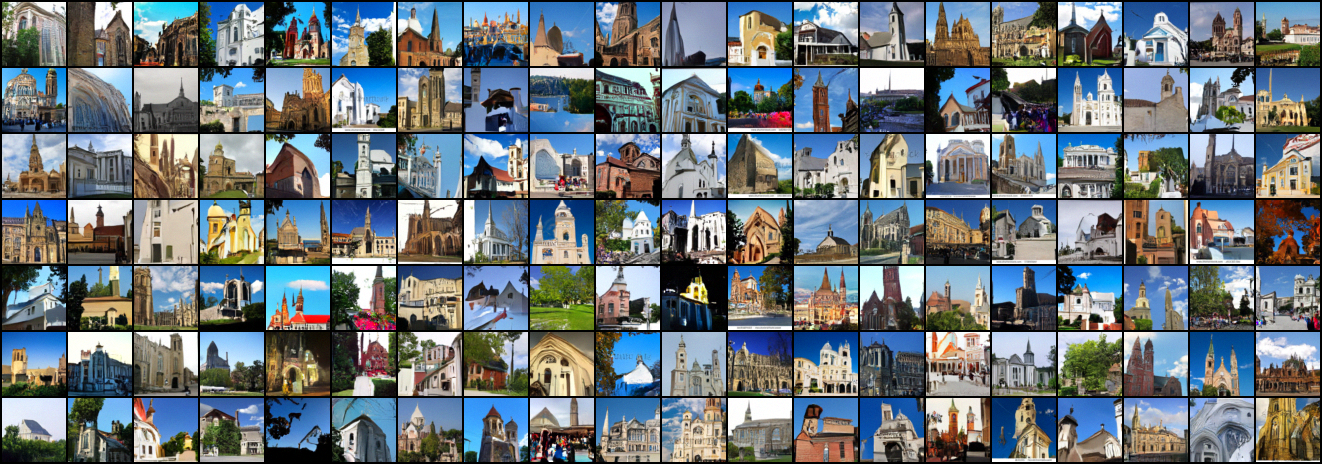}
    \captionof{figure}{Images synthesis on CIFAR-10 and LSUN-Chruch-64.}
    \label{Fig.add-cifar10-syn}
\end{minipage}

\begin{minipage}{\columnwidth}
    \centering
    \includegraphics[width=0.48\columnwidth]{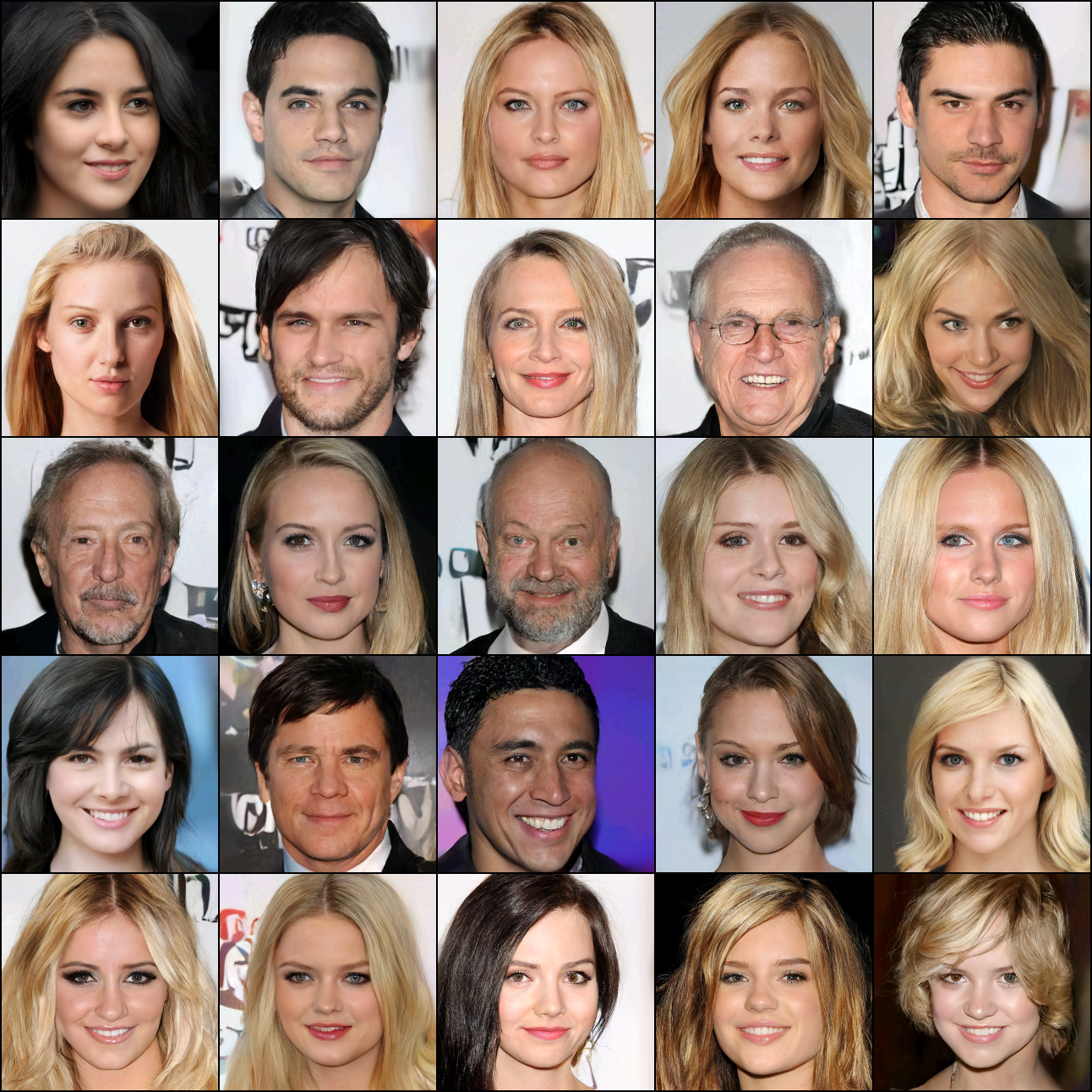}
    \includegraphics[width=0.48\columnwidth]{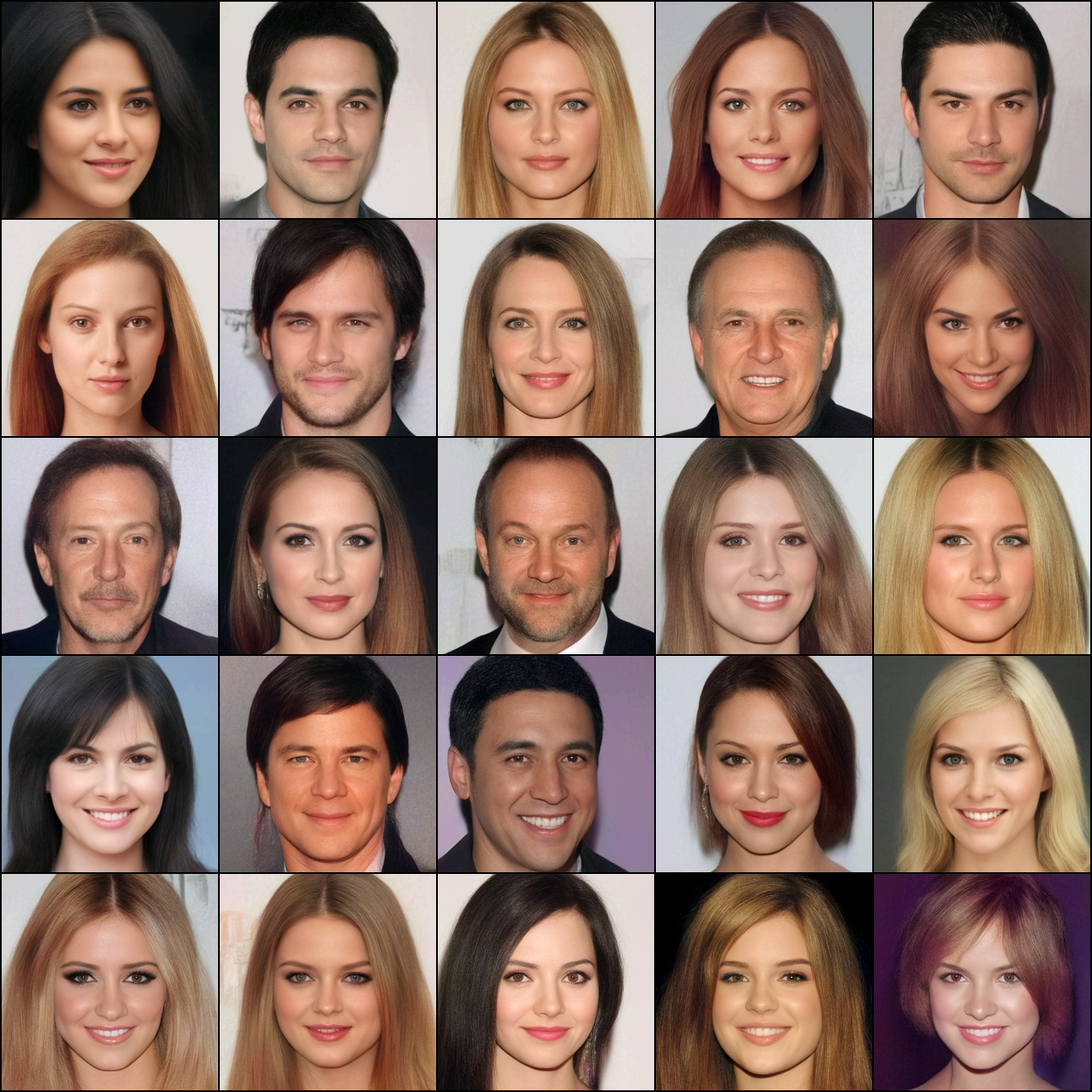}
    \captionof{figure}{Images synthesis on CelebA-HQ-256. Left figure shows temperature=1.0. Right figure shows temperature=0.7.}
    \label{Fig.add-celeba256-syn}
\end{minipage}

\newpage


\end{document}